\documentclass[10pt,twocolumn]{article}
\usepackage{conference}
\usepackage{times}
\usepackage{graphicx}
\usepackage{amssymb}
\usepackage{url,hyperref}
\usepackage{algpseudocode}
\usepackage{algorithm}
\usepackage{subcaption}
\usepackage{amsmath}
\usepackage[round]{natbib} 
\usepackage{stfloats}

\begin{document}

\title{Soft Actor-Critic with Cross-Entropy Policy Optimization}

\author{
	Zhenyang Shi\textsuperscript{1}
	\and
	Surya P. N. Singh\textsuperscript{1}
}

\maketitle

\footnotetext[1]{Robotics Design Laboratory, University of Queensland, Brisbane, Australia. 
Correspondence to: 
Zhenyang Shi \href{mailto:z.shi@uqconnect.edu.au}{$<$z.shi@uqconnect.edu.au$>$},
Surya P.N. Singh \href{mailto:spns@uq.edu.au}{$<$spns@uq.edu.au$>$}.}

\begin{abstract}
Soft Actor-Critic (SAC) is one of the state-of-the-art off-policy reinforcement learning (RL) algorithms that is within the maximum entropy based RL framework. SAC is demonstrated to perform very well in a list of continous control tasks with good stability and robustness. SAC learns a stochastic Gaussian policy that can maximize a trade-off between total expected reward and the policy entropy. To update the policy, SAC minimizes the KL-Divergence between the current policy density and the soft value function density. Reparameterization trick is then used to obtain the approximate gradient of this divergence. In this paper, we propose Soft Actor-Critic with Cross-Entropy Policy Optimization (SAC-CEPO), which uses Cross-Entropy Method (CEM) to optimize the policy network of SAC. The initial idea is to use CEM to iteratively sample the closest distribution towards the soft value function density and uses the resultant distribution as a target to update the policy network. For the purpose of reducing the computational complexity, we also introduce a decoupled policy structure that decouples the Gaussian policy into one policy that learns the mean and one other policy that learns the deviation such that only the mean policy is trained by CEM. We show that this decoupled policy structure does converge to a optimal and we also demonstrate by experiments that SAC-CEPO achieves competitive performance against the original SAC.
\end{abstract}

\section{Introduction}
By recent years of development, model-free deep reinforcement learning (RL) has become capable of learning sophisticated policy in complex environments. Deep Deterministic Policy Gradient (DDPG) is an off-policy Actor-Critic algorithm to solve RL problems with continuous action space \citep{lillicrap2015continuous}. While DDPG is demonstrated to have a fast convergence rate due to its high sampling efficiency, it is sensitive to hyperparamters and lack of stability \citep{duan2016benchmarking, henderson2018deep}. Multiple attempts have been made to combine DDPG with Cross-Entropy Method (CEM) such as Ot-opt \citep{kalashnikov2018qt} which uses CEM to iteratively sample the best policy that maximize the current Q function and the Cross-Entropy Guided Policies (CGP), which, in addition to Ot-opt, trains a deterministic policy from imitating the policy sampled by CEM to improve its inference time efficiency \citep{simmons2019q}.

Soft Actor-Critic (SAC) is also an off-policy Actor-Critic algorithm but it aims to maximize a trade-off between expected total return and the policy entropy \citep{haarnoja2018soft}. Unlike DDPG, SAC is demonstrated to show both efficiency and stability. However, there is yet any attempt to incorporate CEM into SAC. In this paper, we propose Soft Actor-Critic with Cross-Entropy Policy Optimization (SAC-CEPO) that uses CEM to iteratively sample the best policy that maximize the current value function. The sampled policy is then used to train the actual policy network similar to CGP. One major challenge is that DDPG learns a deterministic actor and the policy output is a scalar vector but SAC learns a stochastic actor and the policy output is a vector of Gaussian distributions, which dramatically increases the computational complexity of CEM. We therefore also introduce a decoupled policy structure that decouples the mean and deviation of the ogirinal Gaussian policy from SAC such that CEM is only used to optimize the policy that learns the mean. We prove that this decoupled policy structure with CEM does converge to a local optimal and we also empirically demonstrate that SAC-CEPO can achieve a competitive performance against the ogirinal SAC. Finally, we study the sensetivity of the additional hyperparameters that introduced in SAC-CEPO.

\section{Related Work}
There are a number of reinforcement learning (RL) algorithms based on the Policy Gradient Theorem, which provides an expression for the gradient of a performance measure with respect to the parameters of the policy \citep{sutton2000policy}. REINFORCE or Monte Carlo Policy Gradient is a direct application of the Policy Gradient Theorem but it suffers from high variance and slow learning speed \citep{sutton2000policy}. Trusted Region Policy Optimization (TRPO) uses second order information to update the policy network such that the update is as large as possible with the constraint that the new policy does not differ much from the old policy using a measure of KL-Divergence \citep{schulman2015trust}. Proximal Policy Optimization (PPO) is another variant of REINFORCE that uses first order method to achieve a similar performance as TRPO \citep{schulman2017proximal}. However, REINFORCE, TRPO and PPO are all on-policy methods: They require new experience to be collected for updates of the policy network, making their sampling efficiency low.

Deep Deterministic Policy Gradient (DDPG) is an off-policy RL algorithm based on the Deterministic Policy Gradient Theorem \citep{silver2014deterministic} that has high sampling efficiency \citep{lillicrap2015continuous}. DDPG learns a value function by evaluating the current policy and improve its policy by following the gradient that maximizes the value function in sampled states \citep{lillicrap2015continuous}. Since DDPG learns a deterministic actor, to achieve exploration a noise can be applied to the output of the policy network \citep{lillicrap2015continuous} or to the parameters of the policy network \citep{plappert2017parameter}. Although DDPG is demonstrated to perform well in a range of tasks, it suffers from instability and high sensitivity to hyperparameters \citep{duan2016benchmarking, henderson2018deep}. Ever since then, various changes have been proposed to improve DDPG. Twin Delayed Deep Determinisitc Policy Gradients (TD3) \citep{fujimoto2018addressing} is one of them.

TD3 uses two value functions to address the overestimation bias occured in Actor-Critic \citep{hasselt2010double, fujimoto2018addressing} and a delayed policy update together with a target policy to smooth value function. These changes significantly improve the performance of DDPG. 

Concurrently to TD3, Soft Actor-Critic (SAC) \citep{haarnoja2018soft} is proposed and it is based on the maximum entropy RL framework. In traditional RL such as DDPG, the aim is to find a policy that maximize the total expected reward while in maximum entropy RL, the aim is to find a policy that can maximize a trade-off between the total expected reward and the policy entropy. Besides, DDPG learns a deterministic actor while SAC learns a stochastic actor so the output from the SAC is not a scalar but a Gaussian distribution over action space. Similar to DDPG, SAC learns a soft value function by evaluating the current policy and then improve its policy by minimizing the KL-Divergence between the current policy density and the value function density. More specifically, reparameterization trick is used to obtain the approximate gradient of such KL-Divergence and the gradient can then be used to update the policy network. SAC is demonstrated to outperform DDPG with a substantial margin and shows great stability \citep{haarnoja2018soft}. Our work is based on the SAC algorithm but rather than use the stochastic gradient descent to directly optimize the objective function after the reparameterization trick, we use Cross-Entropy Method (CEM) to optimize the policy network. One possible way is to use CEM to direct iteratively sample a target policy density from the policy space that is closest to the value function density. The sampled policy can therefore be used as a label to update the original policy network.

There are some attempts to use CEM in RL. CEM-RL \citep{pourchot2018cem} combines evolutionary algorithm with RL by using CEM to sample a population of policy networks such that the sampled policy can maximize the value function. Qt-Opt uses RL to learn a value function which is trained using CEM sampled policy and learns to manipulate a real vision based robotics \citep{kalashnikov2018qt}. Cross-Entropy Guided Policies (CGP) is similar to Qt-Opt but trains a deterministic policy to imitate the policy sampled by the CEM to improve its inference time efficiency \citep{simmons2019q}. Our proposed method, Soft Actor-Critic with Cross-Entropy Policy Optimization (SAC-CEPO) is similar to CGP but we trains a stochastic policy to imitate the policy sampled by CEM. 

Ot-Opt and CGP use CEM to sample the action that can maximize the current value function as they train a deterministic policy but SAC-CEPO aims to learn a stochastic policy or more specifically, a Gaussian policy over action space. Therefore, CEM will have to sample a vector of Gaussian distributions iteratively to find the target distribution that can maximize the soft value function. A Gaussian distribution can be represented by a mean and a deviation which means the dimension of the sampled vector is increased by a factor of two compared to sampling a single action. Therefore, it will be much more computational expensive than Ot-Opt or CGP. To reduce the computational complexity, we introduce a decoupled policy network that decouples the Gaussian policy network into one policy network that learns the mean and one other policy network that learns the deviation. CEM will only be performed to sample the best mean that can maximize the current soft value function.

\section{Notation and Background}
We use the notation based in \citep{sutton1998introduction}. Reinforcement Learning (RL) is a subject of Machine Learning that aims to solve Markov Decision Process (MDP). MDP is defined by a tuple of $<S, A, p, r, \gamma>$, where $S$ is the state space, $A$ is the action space, $p$ is the transition probability: $p(s_{t+1}|s_t, a_t)$ is the probability of moving into next state $s_{t+1}$ given the current state $s_t$ and action taken $a_t$, $r$ is the reward function: $r(s_t, a_t, s_{t+1})$ is the reward given for a state transition from $s_t$ to $s_{t+1}$ with action $a_t$, $\gamma $ is the discount factor to weigh future reward. We also use $E_\pi[\cdot]$ to denote the expected value of a random variable if the agent follows policy $\pi$.

In standard RL, the objective is to maximize the total expected reward: $ E_\pi[\sum_{t}\sum_{s'}p(s'|s_t, a_t)r(s_t, a_t, s')]$. However, the objective of Soft Actor-Critic (SAC) is to find a policy $\pi$ that can maximize a trade-off between total expected reward and the entropy of the policy:
\begin{equation}
J(\pi) = E_\pi[\sum_{t}[\sum_{s'}p(s'|s_t, a_t)r(s_t, a_t, s')] + \alpha H(\pi(\cdot|s_t))]
\end{equation}
Where $H(\cdot)$ is the entropy of a given density and $\alpha$ is a temperature parameter to determine the relative importance of the the entropy term against the reward term.

SAC involves a concept of soft value function where the soft $Q$ value and the soft $V$ value is connected through:
\begin{equation}
Q(s_t, a_t) = \sum_{s'}p(s'|s_t, a_t)r(s_t, a_t, s') + \gamma V(s')
\end{equation}
\begin{equation}
V(s_t) = E_{a_t \sim \pi}[Q(s_t, a_t) - log\pi(a_t, s_t)]
\end{equation}

SAC in \citep{haarnoja2018soft} contains five networks: Two soft $Q$ networks parameterized by $\theta_1$ and $\theta_2$ respectively, one soft $V$ network parameterized by $\psi$, one target soft $V$ network parameterized by $\psi'$ and one Gaussian policy network $\pi$ parameterized by $\phi$. 

SAC updates the soft $V$ network to minimize:
\begin{equation}
\resizebox{.95\hsize}{!}{$
J_V(\psi) = E_{s_t \sim D}[\frac{1}{2}(V_\psi(s_t)-E_{a_t \sim \pi_\phi}[Q_\theta(s_t, a_t) - \alpha log\pi_\phi(a_t|s_t)])^2]
$}
\end{equation}
Where $D$ is a replay buffer and $Q_\theta(s_t, a_t)$ is the minimum of two $Q$ networks.
Meanwhile, the $Q$ network is updated by minimizing the soft Bellman error:
\begin{equation}
\label{eq:Q}
\resizebox{.95\hsize}{!}{$
J_Q(\theta) = E_{(s_t, a_t, r_t, s_{t+1}) \sim D}[\frac{1}{2}(Q_\theta(s_t, a_t) - (r_t + \gamma V_{\psi'} (s_{t+1})))^2]
$}
\end{equation}
Finally, SAC updates the policy network by minimizing:
\begin{equation}
J_\pi(\phi) = E_{s_t \sim D}[D_{KL}(\pi_\phi(\cdot|s_t)||\frac{exp(Q_\theta(s_t, \cdot))}{Z_\theta(s_t)})]
\end{equation}
Reparameterization trick is used to convert the objective function into:
\begin{equation}
\resizebox{.95\hsize}{!}{$
J_\pi(\phi) = E_{s_t \sim D, \epsilon_t \sim N}[\alpha log\pi_\phi(f_\phi(\epsilon_t; s_t)|s_t) - Q_\theta(s_t, f_\phi(\epsilon_t;s_t))]
$}
\end{equation}
Where $f_\phi(\epsilon_t; s_t)$ is the reparameterized policy and $\epsilon_t$ is an input noise vector.

Cross-Entropy Method (CEM) is a generic and efficient algorithm to solve rare event probability estimation and stochastic optimization problems \citep{de2005tutorial}. We use a Gaussian distribution in our CEM algorithm and the pseudocode is shown in Algorithm \ref{alg:CEM-ALG}. As a result, there will be four additional parameters introduced: Initial deviation vector (Initial Search Size $s$), Sample Number $N$, Elite Sample Density $\rho_e$ and Number of Iterations $T$. Initial mean vector is the initial guess, which could be any random vector as long as the initial search size is large enough. However, since we choose to run the CEM only for a fixed number of iterations, we set the initial mean vector to be the current prediction from the network.

\begin{algorithm}[t]
\caption{Cross-Entropy Method for optimization with Gaussian distribution}
\begin{algorithmic}[0]
\State \textbf{input:} objective function $f(\theta)$, initial mean vector $\theta_\mu$, initial deviation vector $\theta_\sigma$
\For {$t = 1$ to $T$}
	\State Sample $\theta_1, ..., \theta_N$ from $Normal(\theta_\mu, \theta_\sigma)$
	\State $F = sort(f(\theta_1), ..., f(\theta_N))$
	\State Get elite samples: $F_{elite} = F(1:N \times \rho_e)$
	\State Update mean: $\theta_\mu = \mu(F_{elite})$
	\State Update deviation: $\theta_\sigma = \sigma(F_{elite})$
	\If {$\theta_\sigma < \delta$}
		\State \textbf{break}
	\EndIf
\EndFor
\end{algorithmic}
\label{alg:CEM-ALG}
\end{algorithm}

\section{Soft Actor-Critic (SAC) with decoupled policy network}

Qt-Opt or Cross-Entropy Guided Policies (CGP) uses Cross-Entropy Method (CEM) to sample a deterministic action that can maximize the value function. In SAC, however, the output from the policy network is a unbounded Gaussian distribution over action space \citep{haarnoja2018soft} and using CEM to directly sample a distribution is much more computational expensive than Qt-Opt or CGP. Therefore, we decouple the policy network of SAC into two: $\pi^\mu$ and $\pi^\sigma$ where $\pi^\mu$ learns the mean of the Gaussian distribution while $\pi^\sigma$ learns the deviation. The original policy can be retrived by $\pi = N(\pi^\mu, \pi^\sigma)$. 

In the policy improvement step, for each state, we first update the $\pi^\mu$ according to:
\begin{equation}
\resizebox{.95\hsize}{!}{$
\pi^\mu_{new} = arg\,min_{\pi' }D_{KL}(N(\pi'(s_t), \pi^\sigma_{old}(s_t))||\frac{exp(Q^{N(\pi^\mu_{old}, \pi^\sigma_{old})}(s_t, \cdot))}{Z^{N(\pi^\mu_{old}, \pi^\sigma_{old})(s_t)}})
$}
\end{equation}
After we obtain the $\pi^\mu_{new}$, we then update the $\pi^\sigma$ according to:
\begin{equation}
\resizebox{.95\hsize}{!}{$
\pi^\sigma_{new} = arg\,min_{\pi'}D_{KL}(N(\pi^\mu_{new}(s_t), \pi'(s_t))||\frac{exp(Q^{N(\pi^\mu_{new}, \pi^\sigma_{old})}(s_t, \cdot))}{Z^{N(\pi^\mu_{new}, \pi^\sigma_{old})(s_t)}})
$}
\end{equation}

We can show that new policy $\pi_{new} = N(\pi^\mu_{new}, \pi^\sigma_{new})$ has a higher value than the old policy $\pi_{old} = N(\pi^\mu_{old}, \pi^\sigma_{old})$. 

In fact, for any policy that is a function of a list of subpolicies, we can always improve the overall policy by improving subpolicies in sequence. See Appendix A for proof.

\section{Soft Actor-Critic with Cross-Entropy Policy Optimization (SAC-CEPO)}

Similar to original Soft Actor-Critic (SAC), SAC-CEPO uses two soft state-action network: $Q_{\theta_1}$ and $Q_{\theta_2}$, one soft value network: $V_\psi$, one target soft value network: $V_{\psi'}$, one gaussian policy mean network: $\pi^\mu_\phi$ and one gaussian policy deviation network: $\pi^\sigma_\omega$.

The V network is trained to minimize:
\begin{equation}
\resizebox{.95\hsize}{!}{$
J_V(\psi) = E_{s_t \sim D}[\frac{1}{2}(V_\psi(s_t)-E_{a_t \sim N(\pi^\mu_\phi, \pi^\sigma_\omega)}[Q_\theta(s_t, a_t) - \alpha logN(\pi^\mu_\phi, \pi^\sigma_\omega)(a_t|s_t)])^2]
$}
\end{equation}

Q networks are updated to minimize the same objective function as Equation \ref{eq:Q} as it is not directly related to the policy.

Using reparameterization trick, the gaussian mean policy network is updated to minimize:
\begin{equation}
\label{eq:pi-mu}
\resizebox{1\hsize}{!}{$
J_{\pi^\mu}(\phi) = E_{s_t \sim D, \epsilon_t \sim N}[\alpha logN(\pi^\mu_\phi(s_t), \pi^\sigma_\omega(s_t))(f_{\phi,\omega}(\epsilon_t; s_t)|s_t) - Q_\theta(s_t, f_{\phi,\omega}(\epsilon_t;s_t))]
$}
\end{equation}
Then the gaussian deviation policy network is updated to minimize:
\begin{equation}
\resizebox{1\hsize}{!}{$
J_{\pi^\sigma}(\omega) = E_{s_t \sim D, \epsilon_t \sim N}[\alpha logN(\pi^\mu_{\phi'}(s_t), \pi^\sigma_\omega(s_t))(f_{\phi',\omega}(\epsilon_t; s_t)|s_t) - Q_\theta(s_t, f_{\phi',\omega}(\epsilon_t;s_t))]
$}
\end{equation}
Where $\phi'$ is the updated parameters from Equation \ref{eq:pi-mu}.

Q networks, V network and the gaussian deviation policy network are optimized using stochastic gradient descent while the gaussian mean policy network is optimized indirectly using CEM.

We use CEM to firstly find the approximate best output $\mu_*$ that results in the minimum value of:
\begin{equation}
\resizebox{1\hsize}{!}{$
F(\mu) = E_{s_t \sim D, \epsilon_t \sim N}[\alpha logN(\mu, \pi^\sigma_\omega(s_t))(f_{\mu, \omega}(\epsilon_t;s_t)|s_t) - Q_\theta(s_t, f_{\mu, \omega}(\epsilon_t;s_t))]
$}
\end{equation}

Then, the sampled $\mu_*$ can be used as a target value to update the gaussian mean policy network by minimizing:
\begin{equation}
J'_{\pi^\mu}(\phi)|_{\mu_*} = E_{s_t \sim D}[\frac{1}{2}(\pi^\mu_\phi(s_t) - \mu_*)^2]
\end{equation}

In general, CEM will stop the iteration if the deviation in CEM gets small enough. However in practice, using this stop condition will make the computation time unpredictable due to the noise term in Equation \ref{eq:pi-mu}. SAC-CEPO therefore only run the CEM for a fixed number of iterations. 

In addition, we bound the action to a finite interval by applying an invertible squashing function (tanh) similar to SAC \citep{haarnoja2018soft}. We also take one optimization step after one environment step.

The pseudocode for the complete algorithm is shown in Algorithm \ref{alg:SAC-CEPO-ALG}.

\begin{algorithm}[t]
\caption{Soft Actor-Critic with Cross-Entropy Policy Optimization}
\begin{algorithmic}[0]
\State initialise parameter vectors $\theta_1$, $\theta_2$, $\psi$, $\psi'$, $\phi$, $\omega$
\For {each iteration}
	\For {each environment step}
		\State $a_t \sim N(\pi^\mu_\phi, \pi^\sigma_\omega)$
		\State $s_{t+1} \sim p(\cdot|s_t, a_t)$
		\State $r_t \gets r(s_t, a_t, s_{t+1})$
		\State $D \gets D \cup (s_t, a_t, r_t, s_{t+1})$
		\For {each gradient step}
			\State $\psi \gets \psi - \lambda_V \nabla_\psi J_V(\psi)$
			\State $\theta_i \gets \theta_i - \lambda_Q \nabla_{\theta_i} J_Q(\theta_i)$ for $i \in (1 ,2)$
			\State $\mu_* \gets CEM(F(\mu))$
			\State $\phi \gets \phi - \lambda_{\pi^\mu} \nabla_\phi J'_{\pi^\mu}(\phi)|_{\mu_*}$
			\State $\omega \gets \omega - \lambda_{\pi^\sigma} \nabla_\omega J_{\pi^\sigma}(\omega)$
			\State $\psi' \gets \tau\psi + (1 - \tau)\psi'$
		\EndFor
	\EndFor
\EndFor
\end{algorithmic}
\label{alg:SAC-CEPO-ALG}
\end{algorithm}

\section{Experiment}

We benchmarked our method, Soft Actor-Critic with Cross-Entropy Policy Optimization (SAC-CEPO) with the original Soft Actor-Critic (SAC) and a modified Soft Actor-Critic with only decoupled policy network (SAC-DPN) on a range of continuous control tasks from the OpenAI Gym Benchmark Tool \citep{brockman2016openai}.

We implement our own SAC and make further modifications to implement the SAC-CEPO and the SAC-DPN. We follow the hyperparameters settings in the original SAC and we use the same reward scale to implicitly set the temperature parameter for each task. The hyperparameters between SAC, SAC-CEPO and SAC-DPN are shared except for the additional hyperparameters introduced in the Cross-Entropy Method (CEM). All the hyperparameters settings can be found in Appendix B.

We use five random seeds to train SAC, SAC-DPN and SAC-CEPO for each task. We perform an evaluation of the algorithm every 1000 steps for Hopper-v2 and Walker2d-v2, every 3000 steps for Ant-v2 and HalfCheetah-v2 and every 10000 steps for Humanoid-v2. Each evaluation is an average of 10 rollouts in order to reduce the noise. We also choose a deterministic action during rollouts, that is the action chosen is the mean of the Gaussian distribution rather than a random sample from the distribution. The source code of our implementation of SAC, SAC-CEPO and SAC-DPN is available online\textsuperscript{1}.

\footnotetext[1]{github.com/wcgcyx/SAC-CEPO}

\begin{figure*}[t]
\centering
\subfloat[Hopper-v2]{\includegraphics[width=6cm]{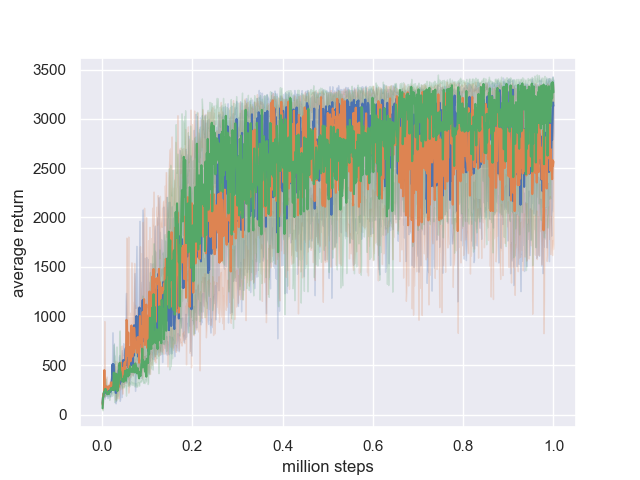}}
\subfloat[Walker2d-v2]{\includegraphics[width=6cm]{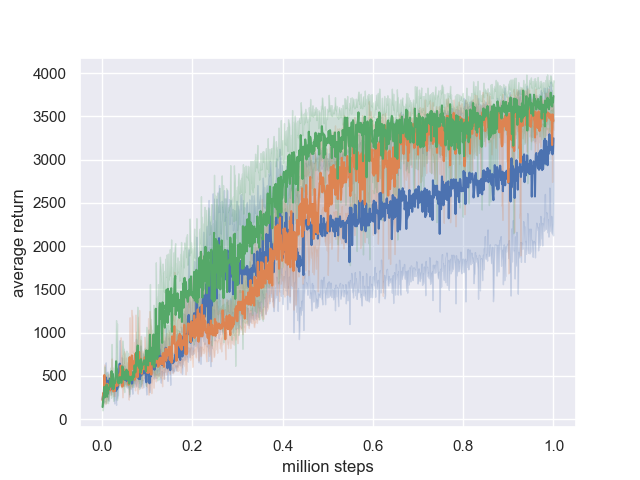}}

\subfloat[Ant-v2]{\includegraphics[width=6cm]{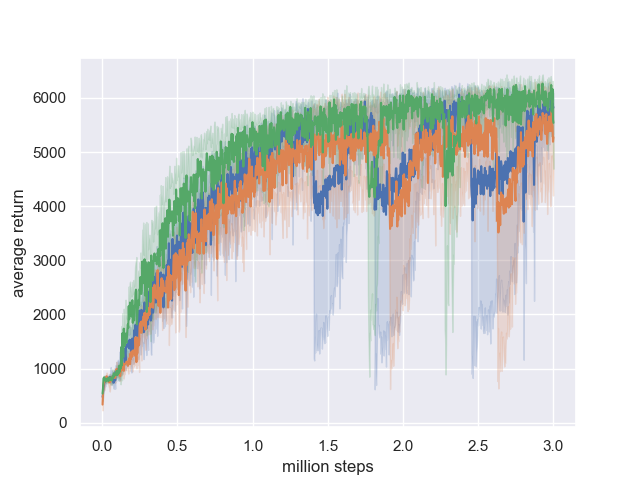}}
\subfloat[HalfCheetah-v2]{\includegraphics[width=6cm]{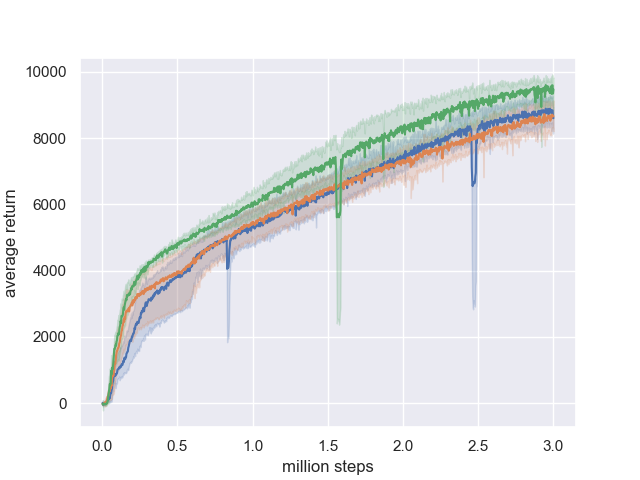}}   
\subfloat[Humanoid-v2]{\includegraphics[width=6cm]{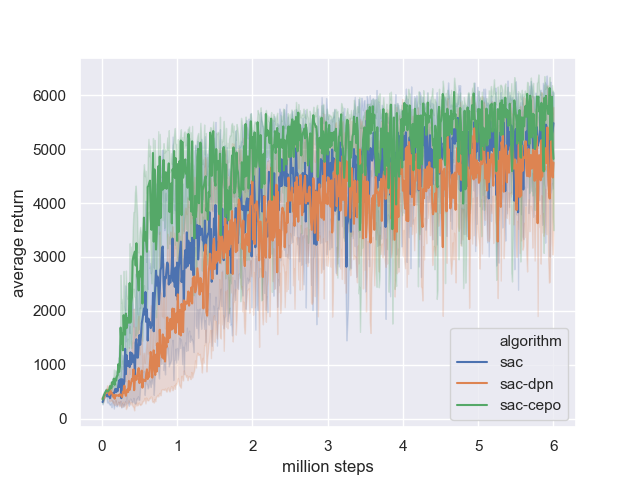}}
\caption{Training curves on continuous control tasks for SAC, SAC-DPN and SAC-CEPO}
\label{fig:BENCHMARK}
\end{figure*}

As shown in Figure \ref{fig:BENCHMARK}, SAC-CEPO achieves a similar performance to SAC-DPN and SAC in Hopper-v2 but outperforms both of them in other tasks especially in complex tasks such as the Humanoid-v2 where SAC-CEPO is able to achieve a reward of 5000 about 3 million steps earlier than SAC and SAC-DPN.

\begin{figure*}[t]
\centering
\subfloat[Sample Number $N$]{\includegraphics[width=6cm]{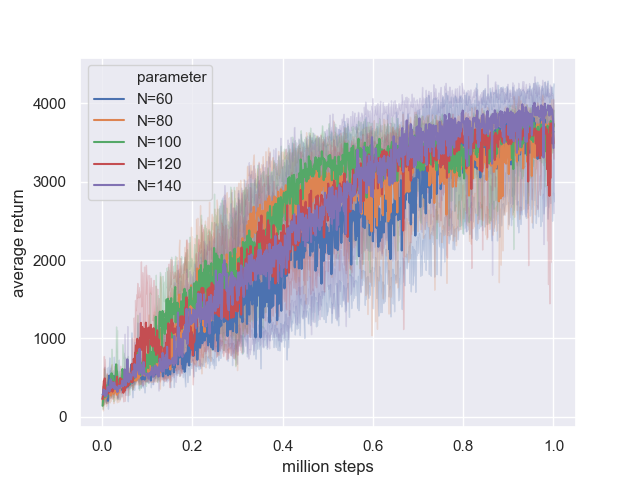}}
\subfloat[Number Of Iterations $T$]{\includegraphics[width=6cm]{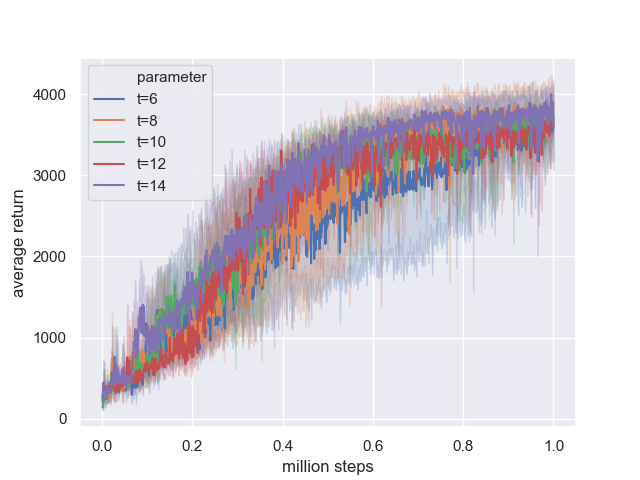}}

\subfloat[Elite Sample Density $\rho$]{\includegraphics[width=6cm]{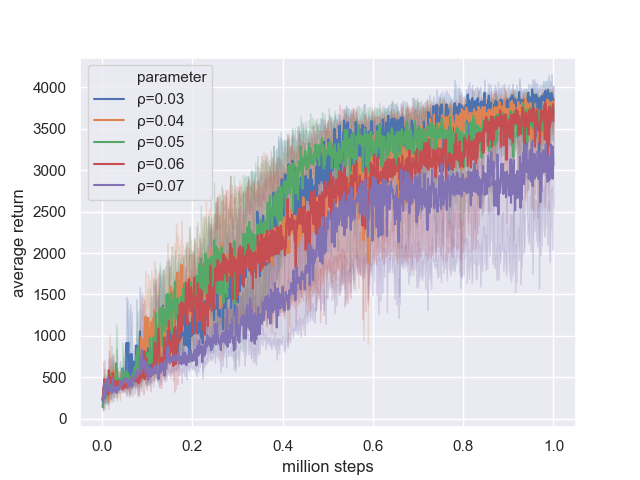}}
\subfloat[Initial Search Size $s$]{\includegraphics[width=6cm]{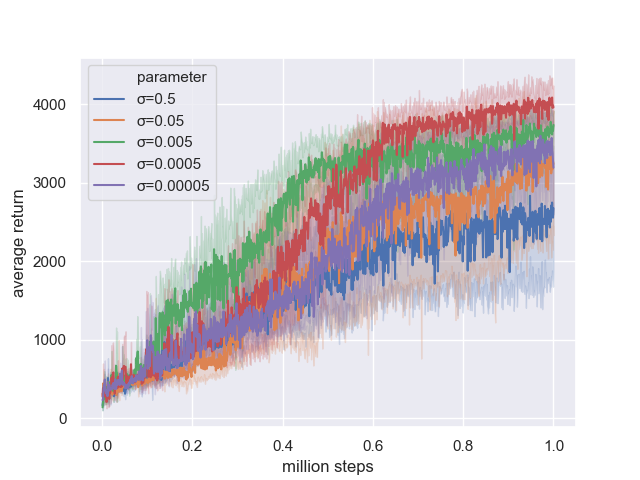}}
\caption{Hyperparameter sensitivity benchmarks for SAC-CEPO on Walker2d-v2}
\label{fig:BENCHMARK-HYPER}
\end{figure*}

We also study the sensitivity of each hyperparameter introduced by CEM by benchmarking on the Walker2d-v2 task. We use the hyperparameters described in Appendix B as a baseline. We conduct one experiment that varies the sample number $N$ from 60 to 140, one experiement that varies number of iterations $T$ from 6 to 14, one experiment that varies the elite sample density $\rho$ from 0.03 to 0.07 and one experiment that varies the initial search size $s$ from 0.5 to 0.00005. The result is shown in \ref{fig:BENCHMARK-HYPER}. 

We found that SAC-CEPO is relatively robust to all four hyperparameters. For sample number $N$, a larger sample number can slightly improve the overall performance but sampling from a large vector is much more computaitonal expensive compared to sampling from a small vector and this inevitably makes each training step slower. Similarly, a larger iteration number $T$ can also increase the overall performance of SAC-CEPO but meanwhile also increase the training time of each step. Elite sample density $\rho$ could affect the performance of SAC-CEPO with a certain degree. In extreme cases, a $\rho$ of 1 is equvalent to random sampling while a $\rho$ of $1/N$ is equivalent to always choose the maximum point from samples. In general, CEM with smaller $\rho$ tends to find a larger value of a given function and therefore SAC-CEPO with smaller $\rho$ usually performs well. However, as $\rho$ becomes very small, CEM is likely to be affected by the noise, which could cause a problem of overfitting. Similarly, under a fixed number of iterations, CEM can find larger value with smaller initial search size $s$ so in general SAC-CEPO with smaller $s$ performs better. However, if $s$ becomes too small, SAC-CEPO is very likely to stuck in a bad local optimal and the performance drops dramatically. For example, when $s<=5e-4$, performance improves with $s$ getting smaller but when $s=5e-5$, the performance is even worse than $s=5e-3$. 

It is worth to notice that SAC-CEPO trains much slower than the original SAC or SAC-DPN due to its nature of iterative sampling. In benchmarking experiment where the number of iterations is set to be 10, SAC-CEPO takes about 10 times longer to train than the original SAC. However, in real world applications where the time of one training step is negligible compared to the time of one environmental step, SAC-CEPO could learn much faster than the original SAC in real time scale as it requires fewer environmental steps.  
\clearpage

\section{Conclusion}

To conclude, we present Soft Actor-Critic with Cross-Entropy Policy Optimization (SAC-CEPO) which incorporates Cross-Entropy Method (CEM) into the original Soft Actor-Critic (SAC). We also introduce a decoupled policy structure that decouples the policy network of SAC into two subpolicies in order to reduce the computational complexity of SAC-CEPO. We empirically show that SAC-CEPO outperforms the original SAC in a range of continuous control tasks. Although one training step in SAC-CEPO takes longer than one training step in SAC, SAC-CEPO could potentially learn much faster than SAC in real world applications where the training time is negligible compared to time involved in interacting with the environment. 

There have been some attempts to replay experience with some prioritity such as DDPG with prioritized experience replay \cite{hou2017novel} and SAC with emphasizing recent experience \cite{wang2019boosting} but neither uses any measure from the actor to prioritize experience. SAC-CEPO introduces a concept of policy error, the distance between sampled policy and the current policy, and such error can be potentially used as a measure of priority in prioritized experience replay. This could be one potential direction for future work.

\clearpage
\bibliographystyle{apalike}
\bibliography{paper}

\begin{thebibliography}{}

\bibitem[Brockman et~al., 2016]{brockman2016openai}
Brockman, G., Cheung, V., Pettersson, L., Schneider, J., Schulman, J., Tang,
  J., and Zaremba, W. (2016).
\newblock Openai gym.
\newblock {\em arXiv preprint arXiv:1606.01540}.

\bibitem[De~Boer et~al., 2005]{de2005tutorial}
De~Boer, P.-T., Kroese, D.~P., Mannor, S., and Rubinstein, R.~Y. (2005).
\newblock A tutorial on the cross-entropy method.
\newblock {\em Annals of operations research}, 134(1):19--67.

\bibitem[Duan et~al., 2016]{duan2016benchmarking}
Duan, Y., Chen, X., Houthooft, R., Schulman, J., and Abbeel, P. (2016).
\newblock Benchmarking deep reinforcement learning for continuous control.
\newblock In {\em International Conference on Machine Learning}, pages
  1329--1338.

\bibitem[Fujimoto et~al., 2018]{fujimoto2018addressing}
Fujimoto, S., Van~Hoof, H., and Meger, D. (2018).
\newblock Addressing function approximation error in actor-critic methods.
\newblock {\em arXiv preprint arXiv:1802.09477}.

\bibitem[Haarnoja et~al., 2018]{haarnoja2018soft}
Haarnoja, T., Zhou, A., Abbeel, P., and Levine, S. (2018).
\newblock Soft actor-critic: Off-policy maximum entropy deep reinforcement
  learning with a stochastic actor.
\newblock {\em arXiv preprint arXiv:1801.01290}.

\bibitem[Hasselt, 2010]{hasselt2010double}
Hasselt, H.~V. (2010).
\newblock Double q-learning.
\newblock In {\em Advances in neural information processing systems}, pages
  2613--2621.

\bibitem[Henderson et~al., 2018]{henderson2018deep}
Henderson, P., Islam, R., Bachman, P., Pineau, J., Precup, D., and Meger, D.
  (2018).
\newblock Deep reinforcement learning that matters.
\newblock In {\em Thirty-Second AAAI Conference on Artificial Intelligence}.

\bibitem[Hou et~al., 2017]{hou2017novel}
Hou, Y., Liu, L., Wei, Q., Xu, X., and Chen, C. (2017).
\newblock A novel ddpg method with prioritized experience replay.
\newblock In {\em 2017 IEEE International Conference on Systems, Man, and
  Cybernetics (SMC)}, pages 316--321. IEEE.

\bibitem[Kalashnikov et~al., 2018]{kalashnikov2018qt}
Kalashnikov, D., Irpan, A., Pastor, P., Ibarz, J., Herzog, A., Jang, E.,
  Quillen, D., Holly, E., Kalakrishnan, M., Vanhoucke, V., et~al. (2018).
\newblock Qt-opt: Scalable deep reinforcement learning for vision-based robotic
  manipulation.
\newblock {\em arXiv preprint arXiv:1806.10293}.

\bibitem[Lillicrap et~al., 2015]{lillicrap2015continuous}
Lillicrap, T.~P., Hunt, J.~J., Pritzel, A., Heess, N., Erez, T., Tassa, Y.,
  Silver, D., and Wierstra, D. (2015).
\newblock Continuous control with deep reinforcement learning.
\newblock {\em arXiv preprint arXiv:1509.02971}.

\bibitem[Plappert et~al., 2017]{plappert2017parameter}
Plappert, M., Houthooft, R., Dhariwal, P., Sidor, S., Chen, R.~Y., Chen, X.,
  Asfour, T., Abbeel, P., and Andrychowicz, M. (2017).
\newblock Parameter space noise for exploration.
\newblock {\em arXiv preprint arXiv:1706.01905}.

\bibitem[Pourchot and Sigaud, 2018]{pourchot2018cem}
Pourchot, A. and Sigaud, O. (2018).
\newblock Cem-rl: Combining evolutionary and gradient-based methods for policy
  search.
\newblock {\em arXiv preprint arXiv:1810.01222}.

\bibitem[Schulman et~al., 2015]{schulman2015trust}
Schulman, J., Levine, S., Abbeel, P., Jordan, M., and Moritz, P. (2015).
\newblock Trust region policy optimization.
\newblock In {\em International conference on machine learning}, pages
  1889--1897.

\bibitem[Schulman et~al., 2017]{schulman2017proximal}
Schulman, J., Wolski, F., Dhariwal, P., Radford, A., and Klimov, O. (2017).
\newblock Proximal policy optimization algorithms.
\newblock {\em arXiv preprint arXiv:1707.06347}.

\bibitem[Silver et~al., 2014]{silver2014deterministic}
Silver, D., Lever, G., Heess, N., Degris, T., Wierstra, D., and Riedmiller, M.
  (2014).
\newblock Deterministic policy gradient algorithms.

\bibitem[Simmons-Edler et~al., 2019]{simmons2019q}
Simmons-Edler, R., Eisner, B., Mitchell, E., Seung, S., and Lee, D. (2019).
\newblock Q-learning for continuous actions with cross-entropy guided policies.
\newblock {\em arXiv preprint arXiv:1903.10605}.

\bibitem[Sutton et~al., 1998]{sutton1998introduction}
Sutton, R.~S., Barto, A.~G., et~al. (1998).
\newblock {\em Introduction to reinforcement learning}, volume 135.
\newblock MIT press Cambridge.

\bibitem[Sutton et~al., 2000]{sutton2000policy}
Sutton, R.~S., McAllester, D.~A., Singh, S.~P., and Mansour, Y. (2000).
\newblock Policy gradient methods for reinforcement learning with function
  approximation.
\newblock In {\em Advances in neural information processing systems}, pages
  1057--1063.

\bibitem[Wang and Ross, 2019]{wang2019boosting}
Wang, C. and Ross, K. (2019).
\newblock Boosting soft actor-critic: Emphasizing recent experience without
  forgetting the past.
\newblock {\em arXiv preprint arXiv:1906.04009}.

\end{thebibliography}

\clearpage
\appendix
\onecolumn

\section{Proof}
Let policy to be a function of $N$ subpolicies: $\pi = F(\pi^1, \pi^2, ..., \pi^N)$ where $\pi^i \in \Pi^i$ and let $Q^{F(\pi^1, \pi^2, ..., \pi^N)}$ and $V^{F(\pi^1, \pi^2, ..., \pi^N)}$ be the corresponding soft state-action value and soft state value.

Define the old policy $\pi_{old}$ be $F(\pi^1_{old}, \pi^2_{old}, ..., \pi^N_{old})$ and define the $i^{th}$ new policy $\pi_{new[i]}$ to be:

\begin{equation}
F(\pi^1_{new}, ..., \pi^{i-1}_{new}, \pi^{i}_{new}, \pi^{i+1}_{old}, ..., \pi^N_{old})
\end{equation}

Which is further defined to be:

\begin{equation}
\begin{aligned}
\pi_{new[i]}(\cdot|s_t) &= arg\,min_{\pi' \in \Pi^i} D_{KL}(
F(..., \pi^{i - 1}_{new}, \pi', \pi^{i + 1}_{old}...)(\cdot|s_t)||\frac{exp(Q^{\pi_{new[i-1]}}(s_t, \cdot))}{log Z^{\pi_{new[i-1]}}(s_t)}) \\
&= arg\,min_{\pi' \in \Pi^i} J_{\pi_{new[i-1]}}(F(..., \pi^{i - 1}_{new}, \pi', \pi^{i + 1}_{old}...)(\cdot|s_t))
\end{aligned}
\end{equation}

It is clear that $J_{\pi_{new[i-1]}}(\pi_{new[i]}) \leq J_{\pi_{new[i-1]}}(\pi_{new[i-1]})$ as we could choose $\pi^i_{new} = \pi^i_{old} \in \Pi^i$. Therefore, we have:

\begin{equation}
\label{eq:bound}
\begin{aligned}
&E_{a_t \sim \pi_{new[i]}}[log\pi_{new[i]}(a_t|s_t) - Q^{\pi_{new[i-1]}}(s_t, a_t) + log Z^{\pi_{new[i-1]}}(s_t)] \\
&\leq  E_{a_t \sim \pi_{new[i-1]}}[log\pi_{new[i-1]}(a_t|s_t) - Q^{\pi_{new[i-1]}}(s_t, a_t) + log Z^{\pi_{new[i-1]}}(s_t)]\\
& \implies E_{a_t \sim \pi_{new[i]}}[Q^{\pi_{new[i-1]}}(s_t, a_t) - log \pi_{new[i]}(a_t|s_t)] \geq V^{\pi_{new[i-1]}}(s_t)
\end{aligned}
\end{equation}

Also:

\begin{equation}
\begin{aligned}
Q^{\pi_{new[i-1]}}(s_t, a_t) &= \sum_{s'} [p(s'|s_t, a_t)r(s_t, a_t, s') + \gamma V^{\pi_{new[i-1]}}(s')] \\
&\leq \sum_{s'} [p(s'|s_t, a_t)r(s_t, a_t, s') + \gamma E_{a_{t+1} \sim \pi_{new[i]}}[Q^{\pi_{new[i-1]}}(s_{t+1}, a_{t+1}) - log \pi_{new[i]}(a_{t+1}|s_{t+1})]] \\
&... \\
&\leq Q^{\pi_{new[i]}}(s_t, a_t)
\end{aligned}
\end{equation}

The soft Bellman equation and Equation \ref{eq:bound} is applied to the right hand side repeatedly. Convergence to $Q^{\pi_{new[i]}}$ can be proved by the Soft Policy Evaluation Lemma \citep{haarnoja2018soft}.

Therefore, we can easily show that:

\begin{equation}
\begin{aligned}
Q^{\pi_{old}}(s_t, a_t) &\leq Q^{\pi_{new[1]}}(s_t, a_t) \\
&\leq Q^{\pi_{new[2]}}(s_t, a_t) \\
&... \\
&\leq Q^{\pi_{new[N]}}(s_t, a_t) = Q^{\pi_{new}}(s_t, a_t)
\end{aligned}
\end{equation}

\clearpage
\section{Hyperparameters}
Table \ref{tab:hyper-param} lists the hyperparameters shared by SAC and SAC-CEPO. Table \ref{tab:scale} lists the shared reward scale for each task. Table \ref{tab:cem-param} lists the additional CEM hyperparameters used in SAC-CEPO for each task.

\begin{table}[h]
\centering
\begin{tabular}{l|l}
\hline
Parameter                              & Value     \\ \hline
Optimizer                              & Adam      \\
Learning rate                          & 0.0003    \\
Discount                               & 0.99      \\
Replay buffer size                     & 1,000,000 \\
Number of hidden layers (all networks) & 2         \\
Number of hidden units per layer       & 256       \\
Number of samples per minibatch        & 256       \\
Activation function                    & ReLU      \\
Target smoothing coefficient           & 0.005     \\ \hline
\end{tabular}
\caption{Hyperparameters shared by SAC and SAC-CEPO}
\label{tab:hyper-param}
\end{table}

\begin{table}[h]
\centering
\begin{tabular}{llll}
\hline
Task name       & State Space Dimension & Action Space Dimension & Reward Scale \\ \hline
Hopper-v2       & 11                    & 3                      & 5            \\
Walker2d-v2     & 17                    & 6                      & 5            \\
HalfCheetach-v2 & 17                    & 6                      & 5            \\
Ant-v2          & 111                   & 8                      & 5            \\
Humanoid-v2     & 376                   & 17                     & 20           \\ \hline
\end{tabular}
\caption{Reward Scale shared by SAC and SAC-CEPO}
\label{tab:scale}
\end{table}

\begin{table}[h]
\centering
\begin{tabular}{lllll}
\hline
Task name       & CEM-Size    & CEM-N      & CEM-       & CEM-Iteration \\ \hline
Hopper-v2       & 0.005        & 100        & 5\%        & 10            \\
Walker2d-v2     & 0.005        & 100        & 5\%        & 10            \\
HalfCheetach-v2 & 0.005        & 100        & 5\%        & 10            \\
Ant-v2          & 0.005 & 100 & 5\% & 10     \\
Humanoid-v2     & 0.01 & 100 & 5\% & 20     \\ \hline
\end{tabular}
\caption{CEM Parameters in SAC-CEPO}
\label{tab:cem-param}
\end{table}

\end{document}